\documentclass[11pt]{article}

\usepackage[preprint]{acl}

\usepackage{times}
\usepackage{latexsym}

\usepackage[T1]{fontenc}

\usepackage[utf8]{inputenc}

\usepackage{microtype}

\usepackage{inconsolata}

\usepackage{graphicx}

\usepackage{hyperref}
\usepackage{url}
\usepackage{microtype}
\usepackage{hyperref}
\usepackage{url}
\usepackage{booktabs}
\usepackage{graphicx}
\usepackage{newtxtext}
\usepackage{amssymb}
\usepackage{longtable}
\usepackage{textcomp}
\usepackage{caption}
\usepackage{latexsym}
\usepackage[most]{tcolorbox}
\usepackage{booktabs}
\usepackage{subcaption}
\usepackage{float}
\usepackage{algpseudocode}
\usepackage{algorithm}
\usepackage{multirow}
\usepackage{comment}
\usepackage{multicol}
\usepackage{adjustbox}
\usepackage{graphicx}
\usepackage{placeins}
\usepackage[T1]{fontenc}
\usepackage{array}

\title{From Personas to Plot: Character-Grounded Multi-Agent Story Generation for Long-Form Narratives}

\author{
  \textbf{Chloe Ho}\textsuperscript{1,*},
  \textbf{Aayush Aluru}\textsuperscript{*},
  \textbf{Muhammad Hammouri}\textsuperscript{2},
  \textbf{Kerry Luo}\textsuperscript{3}
\\[1mm]
  \textbf{Ryan Lagasse}\textsuperscript{$\dagger$},
  \textbf{Arjun Bahuguna}\textsuperscript{4,$\dagger$},
  \textbf{Vasu Sharma}\textsuperscript{$\dagger$}
\\[1mm]
  Pocket FM
\\[1mm]
  \texttt{ch4941@princeton.edu, aayush.aluru09@gmail.com}
}

\newcommand{\magnet}{\textsc{Magnet }}

\begin{document}
\maketitle

\begingroup
\renewcommand{\thefootnote}{1}
\footnotetext{Princeton University.}
\renewcommand{\thefootnote}{2}
\footnotetext{University of Michigan.}
\renewcommand{\thefootnote}{3}
\footnotetext{University of Maryland.}
\renewcommand{\thefootnote}{4}
\footnotetext{Universitat Pompeu Fabra.}
\renewcommand{\thefootnote}{*}
\footnotetext{Equal contribution.}
\renewcommand{\thefootnote}{\ensuremath{\dagger}}
\footnotetext{Senior author.}
\endgroup
\begin{abstract}
Although large language models (LLMs) have demonstrated impressive creative fiction generation, they struggle to maintain narrative consistency and coherent plot lines in long-form stories. In this work, we introduce \textsc{Magnet}, a multi-agent goal-driven narrative engine for storytelling, which generates stories with persona-grounded character agents that propose actions based on a shared structured state and evolving story goals. We evaluate \magnet on 20 and 100 page stories using LLM based editor annotations and rubric scoring. At both narrative lengths, \magnet significantly reduces editor annotations and improves rubric scores compared to single-model prompting, StoryBox, and StoryWriter ($p<0.05$). Ablation experiments also show that the critic module, structured state, and goal sequencing each contribute significantly to performance ($p<0.05$). These results suggest that long-form narratives can emerge from explicit structured states, critic-based action refinement, and goal-driven multi-agent generation with loosely specified character personas, providing a foundation for controllable and structurally coherent long-form narrative generation. 
\end{abstract}
\FloatBarrier
\section{Introduction}

Large language models (LLMs) have significantly advanced open-ended text generation, developing coherent stories and producing systems that can role-play storybook characters and simulate complex interactions \cite{wang2024rolellmbenchmarkingelicitingenhancing,openai2024gpt4technicalreport}. Although LLMs have demonstrated strong narrative generation capabilities, they suffer from character inconsistency and plot discontinuity, limiting their ability to create high-quality long-form narratives \cite{lu2026assistantaxissituatingstabilizing,shao2023characterllmtrainableagentroleplaying,yao2019planandwritebetterautomaticstorytelling}. 

These failures become pronounced in multi-character environments, where LLMs struggle to balance narrative goals and character actions with complex relationships and interactions \cite{park2023generativeagentsinteractivesimulacra,gao2024agentscopeflexiblerobustmultiagent,li2026loststoriesconsistencybugs}. Recent work, including StoryBox \cite{Chen_2026}, Agents' Room \cite{huot2025agentsroomnarrativegeneration}, and StoryWriter \cite{xia2025storywritermultiagentframeworklong} have explored the use of multi-agent systems in narrative generation where specialized agents collaborate to improve coherence and mitigate hallucinations, but they continue to rely on textual memory, limiting their ability to generate coherent stories. \cite{lewis2021retrievalaugmentedgenerationknowledgeintensivenlp,shinn2023reflexionlanguageagentsverbal}. As a result, the story becomes incoherent and deviates from established motivations \cite{liu2026narrativetheorydrivenllmmethods,surveyonllmsforstorygeneration}.

In this work, we introduce \textbf{\textsc{Magnet}}, a multi-agent story generation system to develop coherent long-form narratives. Our work aims to address the research question: \textbf{Can long-form narratives emerge from interactions between character personas and updating goal states when autonomous agents interact through a shared structured state?} Through hierarchical editorial evaluation and rubric scoring analysis, we show that through iterative character-driven actions, goal sequencing, and a structured world state
\magnet improves narrative coherence, providing a foundation for future work in long-form narrative generation. 

Our main contributions include: \newline
\setlength{\parindent}{1.5em} 
\indent 1) \textbf{Character and goal driven narrative generation} that grounds character agents in individual personas with goals \newline
\indent 2) \textbf{A multi-agent Action-Critic-Narrator-Goal interaction loop} that integrates character-grounded action generation, critic revision, narrator-driven prose writing, and evolving story goals for long-range coherence \newline
\indent 3) \textbf{Empirical evaluation metrics} showing that \magnet reduces editorial critique counts and increases rubric scores in long-form stories \newline

\FloatBarrier
\vspace{-1 em}
\section{Related Work}
\textbf{Multi-Agent Narrative Generation} Recent work has demonstrated that the decomposition of narrative generation into interacting language model agents, where each agent is responsible for a high-level role, such as planning, character development, or scene writing, can improve long-form narrative coherence \cite{xia2025storywritermultiagentframeworklong,huot2025agentsroomnarrativegeneration}. Additionally, frameworks such as IBSEN \cite{han2024ibsendirectoractoragentcollaboration} and StoryVerse \cite{Wang_2024} introduce director-actor architectures where a planning module guides character agents through structured narrative objectives, showing that separating global narrative planning from local character behavior allows for consistency and controllability \cite{wu2023autogenenablingnextgenllm,borawski2026worldgenquestlinedependencydrivenprompt}. However, these systems still rely heavily on prompt-level coordination, where narrative state is maintained in text rather than tracked through a persistent shared representation \cite{packer2024memgptllmsoperatingsystems}. Conversely, our approach focuses on narrative generation from loosely-defined character personas and story goals, where coherence is maintained through a shared evolving structured state and character agent-driven action generation. 

\textbf{Structured State Tracking} A key challenge in long-form narrative generation is maintaining consistency over time, especially in multi-character environments \cite{xia2025storywritermultiagentframeworklong}. Prior work in structured generation has explored the use of explicit state representations to improve consistency in text-based environments \cite{huang2026stateconsistencybehaviorconsistency,li2026wordworldlargelanguage}. For example, work on instilling commonsense knowledge into interactive agents 
has demonstrated that structured representations of world states can improve action prediction and narrative coherence in decision-making tasks \cite{chi2025wowworldomniscientworld,ding2025understandingworldpredictingfuture,chen2024llmstateopenworldstate}. Similarly, benchmarks such as STAGE \cite{tian2026stagefullscreenplaybenchmarkreasoning} have highlighted the importance of representing entities, events, and relationships explicitly for evaluating narrative understanding and generation. Other work in text-based game environments, such as TextWorld \cite{cote2019textworldlearningenvironmenttextbased}, further show that agents benefit from maintaining structured representations of environment states in both reinforcement and unsupervised methods that operate for long durations. Our work builds upon these ideas by incorporating a structured state representation, allowing narrative facts, character states, and sequentially-updating goals to evolve across scenes without requiring external supervision. 

\textbf{Story Evaluation} Evaluating long-form generated content remains an open problem \cite{wu2025longevalcomprehensiveanalysislongtext}, particularly for character and environment consistency, plot coherence, and story engagement. Recent work has shown that LLM-as-a-Judge methods can achieve substantial agreement with human preferences on open-ended generation tasks, making LLM-as-a-Judge a scalable alternative to costly human evaluation \cite{zheng2023judgingllmasajudgemtbenchchatbot,liu2023gevalnlgevaluationusing,gu2025surveyllmasajudge}. Structured evaluation frameworks have further demonstrated that explicit criteria and rubrics can improve alignment with human judgments \cite{kim2024prometheusinducingfinegrainedevaluation,rao2026autorubricunifyingrubricbasedllm}. For long-form generation specifically, hierarchical evaluation can assess quality at multiple levels of granularity and achieve strong alignment with human evaluation \cite{que2024hellobenchevaluatinglongtext}. 


\section{Methodology}
\begin{figure}[h]
    \centering 
    {\includegraphics[width=0.8\linewidth]{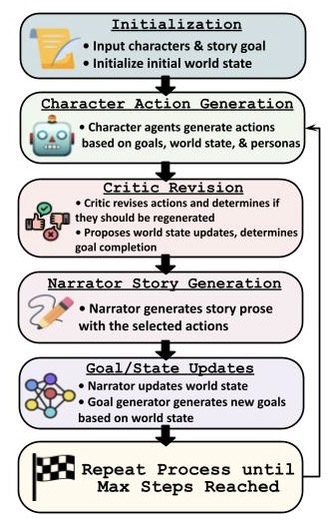}}
    \vspace{-1em}
    \caption{\textsc{Magnet's} generation pipeline. }
    \label{fig:1}
\end{figure}

Our framework creates long-form story generation through iterative multi-agent generation. Instead of directly generating stories from a single prompt, we split generation into three stages: character action generation, critic-based action revision, and narration. At each time step, character agents independently propose actions that are evaluated by a critic module. The narrator then uses select actions to create story prose. The generated paragraph and changes to the structured state are then saved for the next timestep. Story progression is guided by a dynamic goal-generation system that is updated as the story evolves, allowing for new objectives and reducing stalled progression. 

\FloatBarrier
\subsection{Goal Sequencing}
Each story begins with a high-level goal that provides direction for narrative development. When the active goal is completed, an Opus 4.7 \cite{opus4.7} goal generator generates a follow-up goal. Empirically, we observe that after roughly 15 time steps, character actions became increasingly repetitive. To avoid stalled narratives, our framework replaces goals that have not been completed in 15 time steps. After approximately 40 steps, successive goals also began to repeat similar story trajectories. To maintain narrative diversity, the system generates a goal that changes the direction of the story into a new domain (i.e., from a professional conflict into a family conflict) every 40 steps. These larger transitions are intended to introduce new conflicts and shift the story in a new direction as the previous goal concludes. 

\FloatBarrier
\subsection{Story Generation}

Each character is defined with a character persona that contains attributes such as relationships, personality, goals, character description, role, location, and abilities. At each time step, character agents receive the character persona, current story goal, recent story history, and prior world variables that are relevant to that character, and use Gemma-4-31B-it model \cite{gemma4} to generate an action that describes what the character intends to do next. 

After an action has been generated, a Gemini 2.5 Flash \cite{comanici2025gemini25pushingfrontier} LLM critic is asked to evaluate the quality of the proposed action. The critic is provided with the action, character persona, current story goal, and current structured state, and is instructed to determine whether the action is relevant, specific, and consistent with both the character and the current scene. If an action does not effectively advance the goal of the story, is vague, out of character, implausible, or repetitive, the critic will provide feedback for the character agent to generate another action. If the action is accepted or the number of max revisions is reached, the critic generates updates to the world state related to the proposed action.  

After all characters have generated an action, an Opus 4.7 \cite{opus4.7} narrator produces the next paragraph of the story. The narrator receives the proposed actions, current world state, current story goal, and prior story paragraphs. Then it selectively chooses actions that are best suited for the current scene and creates coherent story prose using the selected actions. 

If the narrator selects an action that the critic deemed completed the previous goal, the Opus 4.7 \cite{opus4.7} goal generator will generate a new goal. When generating a new goal, the model receives the previous goal, recent story paragraphs, current world state, all character personas, and all prior goals. Using this context, the goal generator is instructed to produce a new concrete goal that is relevant, not repetitive, and will allow the story to develop further. In addition to generating the next goal, the model also generates a transition paragraph that bridges the previous story paragraph into the new goal. The transition is written as story prose rather than state updates or summary text and is included in the final story.

\begin{center}
\hrule height 1pt
\vspace{2pt}
\captionof{algorithm}{Story generation for one timestep}
\label{alg:one_timestep}
\vspace{-7pt}
\hrule
\vspace{2pt}
\small
\begin{algorithmic}[1]
\Require 
Structured state $W$, goal $g$, characters $C$, stall count $k$, timestep $t$, recent story paragraphs $s$

\vspace{1em}
\Statex \textbf{Action Generation + Critic Revision}
\ForAll{$c \in C$}
    \Repeat
        \State $a \gets \textsc{CharacterAgent}(c, g, W_c, \textit{s})$
        \State $r \gets \textsc{Critic}(a, c, g, W)$
    \Until{$r.\textit{revise} = \texttt{False}$ or MAX\_REVISIONS}
    \State $\textit{actions}[c] \gets (a,\ r.\textit{world\_updates},\ r.\textit{goal\_reached})$
\EndFor

\vspace{1em}
\Statex \textbf{Narrator}
\State $\textit{selected} \gets
\textsc{Narrator}(\textit{actions}, g, W, \textit{s})$
\State append $\textit{selected}.\textit{paragraph}$ to story

\vspace{1em}
\Statex \textbf{World-State Commit}
\State $\textit{goal\_reached} \gets \texttt{False}$

\ForAll{$c \in \textit{selected}.\textit{order}$}
    \ForAll{$(key, val) \in
    \textit{actions}[c].\textit{world\_updates}$}

        \State $W[key] \gets val$
    \EndFor

    \If{$\textit{actions}[c].\textit{goal\_reached}$}
        \State $\textit{goal\_reached} \gets \texttt{True}$
    \EndIf
\EndFor

\vspace{1em}
\Statex \textbf{Goal Sequencing}
\If{$t \bmod 40 = 0$}
    \State $g \gets
    \textsc{GoalGenerator}(W, g,\ \texttt{domain\_shift=True})$
    \State $k \gets 0$
\ElsIf{$k \ge 15$}
    \State $g \gets
    \textsc{GoalGenerator}(W, g,\ \texttt{status="stalled"})$
    \State $k \gets 0$
\ElsIf{$\textit{goal\_reached}$}
    \State $g \gets
    \textsc{GoalGenerator}(W, g,\ \texttt{status="complete"})$
    \State $k \gets 0$
\Else
    \State $k \gets k + 1$
\EndIf
\vspace{2pt}
\hrule
\end{algorithmic}
\end{center}

\FloatBarrier
\subsection{Structured State}
We represent the story state as a directed graph that contains a root world node, character nodes, state variable nodes, and edges that represent the relationships between the nodes. State variable nodes are flattened into a dictionary representation that is provided to the agents during generation. To update the structured state, the selected actions' state updates are applied using an overwrite conflict resolution strategy. If an update writes to a previously unseen key, that key is inserted; if it writes to an existing key, the new value replaces the old one. For example, Lucia begins with holding\_back\_evidence=true, but later, after she reveals the withheld letter, the committed graph contains holding\_back\_evidence=false and lucia\_has\_disclosed\_letter=true. We apply these updates sequentially, but only after narrator selection, so unselected actions' updates do not enter the structured state. If a critic proposes an implausible update, it is typically filtered out because the associated action is not selected. If multiple characters have conflicting updates, the narrator usually only selects the subset of actions that do not conflict, and only those actions' updates are merged. In practice, this keeps overwrite simple while limiting new inconsistencies. 

\FloatBarrier
\subsection{Editorial and Rubric-based Evaluation}
\label{sec:eval pipeline}
We use two LLMs, GPT 5.4 mini \cite{gpt5.4mini} and Claude Haiku 4.5 \cite{haiku4.5}, as expert story editors to analyze generated stories at multiple narrative levels. The LLM evaluators are asked to produce editorial annotations on the whole story, 5 random sections, and 5 random sentences. At the story level, the evaluator analyzes logical consistency, thematic coherence, and character arc completion. At the chapter level, the evaluator analyzes goal-conflict-outcome progression, hook-and-close quality, and chapter necessity. At the sentence level, the evaluator critiques rhythm, clarity, and syntax variety. This hierarchical evaluation framework is designed to evaluate both high-level narrative coherence and lower-level writing quality. For each story, we sum the number of annotations produced by each evaluator across all categories and narrative levels, and then average these totals across the two evaluators to obtain the final annotation count for each story. 

Let $e \in {1,2}$ index the two LLM evaluators, and let $A_{e,l}^{(i)}$ denote the total number of editorial annotations produced by evaluator $e$ for story $i$ at narrative level $l \in {S,C,T}$, corresponding to the story, chapter, and sentence levels, respectively. The final annotation count for story $i$ is computed by summing annotations across the three narrative levels and averaging across evaluators:

\begin{equation}
A^{(i)}
=
\frac{1}{2}
\sum_{e=1}^{2} (A_{e,S}^{(i)} + A_{e,C}^{(i)} + A_{e,T}^{(i)}).
\end{equation}

Additionally, we perform rubric-based scoring evaluation with the same two LLM evaluators. Each evaluator is provided with a story and is instructed to evaluate the story in all categories across the three hierarchies. The evaluator assigns a rubric score out of 100 for each category and also assigns an overall quality score out of 100 for each hierarchy level. For each evaluator, we average the overall quality scores across the story, chapter, and sentence levels and then average these scores across the three evaluators to obtain the final score for each story. 

Let $e \in {1,2}$ index the two LLM evaluators and $\ell \in {S,C,T}$ denote the story, chapter, and sentence level evaluations, respectively. Let $R_{e,\ell}^{(i)}$ denote the overall quality score assigned by evaluator $e$ to story $i$ at narrative level $\ell$. We first compute the mean score assigned by each evaluator across the three hierarchy levels as
\begin{equation}
R_e^{(i)} =
\frac{1}{3}
\left(
R_{e,s}^{(i)} +
R_{e,c}^{(i)} +
R_{e,t}^{(i)}
\right).
\end{equation}
The final rubric-based score for story $i$ is then computed by averaging across the two evaluators: 
\begin{equation}
R^{(i)} =
\frac{1}{2}\sum_{e=1}^{2} R_e^{(i)}.
\end{equation}

\FloatBarrier
\section{Experiments and Results}
\subsection{Experimental Setup}
We evaluate \magnet at generation lengths of 20 (around 10-15k words) and 100 pages (around 50-60k words). For each length, we evaluate five different story generations and report the average results. Before running the system, the user needs to input both a high-level goal that the story aims to achieve and character definitions for each character involved. 

\FloatBarrier
\subsection{Baselines}
\label{sec:baseline}
We compare \magnet against three baselines: a standard prompting approach where Opus 4.7 \cite{opus4.7} generates a story without explicit character agents, a critic module, a structured state, or goal sequencing, StoryBox \cite{Chen_2026}, and StoryWriter \cite{xia2025storywritermultiagentframeworklong}. Unlike the single pass baseline, Storybox and StoryWriter are multi-agent generation frameworks. In StoryBox, story generation is guided by a centralized storyteller who determines the narrative progression  by sequentially deriving the plot goals from the events produced by simulating interactions between independently acting personas. In StoryWriter, story generation is guided by a fixed event outline drafted from the premise, then subdivided into a chapter plan whose sub-events are handed one at a time to a story writing agent. We selected these baselines because they provide publicly available implementations of multi-agent storytelling frameworks, enabling reproducible comparisons. To keep the comparison fair, we provide the baselines with the same story definitions used for \textsc{Magnet}. 

\subsection{Performance Analysis}
\magnet consistently achieves the lowest annotation counts across both generation lengths (Table~\ref{tab:main_results}). Beyond improvements in average annotations counts, paired t-tests demonstrate that \magnet significantly outperforms every baseline at both generation lengths. At 20-page stories, comparisons against the single model, StoryBox, and StoryWriter yield $p = 0.0002$, $p = 0.0016$, and $p = 0.0020$, respectively. This trend persists at 100 pages with $p = 0.0001$, $p = 0.0382$, and $p = 0.0108$, respectively. 

The rubric-based scoring evaluation further demonstrates \textsc{Magnet's} improvements, with \magnet significantly outperforming all three baselines at both generation lengths (Table~\ref{tab:rubric_results}). At 20 pages, paired t-tests against the single model, StoryBox, and StoryWriter yield $p = 0.0043$, $p = 0.0042$, and $p = 0.0206$, respectively. At 100 pages, the improvements remain statistically significant with $p = 0.00033$, $p = 0.0333$, and $p = 0.0424$, respectively. 

\begin{table}[H]
\centering
\setlength{\tabcolsep}{4pt}
\renewcommand{\arraystretch}{0.95}

\begin{tabular}{@{}lcc@{}}
\toprule
\textbf{Method} & \textbf{20 Pages} $\downarrow$ & \textbf{100 Pages} $\downarrow$ \\
\midrule
Single Model & 208.0 $\pm$ 22.02 & 200.7 $\pm$ 10.59 \\
StoryBox & 104.9 $\pm$ 16.18 & 105.4 $\pm$ 9.56 \\
StoryWriter & 120.3 $\pm$ 13.13 & 106.4 $\pm$ 11.52 \\
\textbf{\magnet} & \textbf{84.0} $\pm$ \textbf{9.64} & \textbf{86.4} $\pm$ \textbf{7.80} \\
\bottomrule
\end{tabular}

\caption{Editorial annotation counts averaged across five generated stories.
Lower values indicate fewer critiques. Detailed story length breakdowns
can be found in~\ref{app:editor annotation breakdown}.}
\label{tab:main_results}
\end{table}

\begin{table}[H]
\centering
\setlength{\tabcolsep}{5pt}
\renewcommand{\arraystretch}{0.95}

\begin{tabular}{@{}lcc@{}}
\toprule
\textbf{Method} & \textbf{20 Pages} $\uparrow$ & \textbf{100 Pages} $\uparrow$ \\
\midrule
Single Model & 76.8 $\pm$ 3.71 & 71.0 $\pm$ 2.95 \\
StoryBox & 76.0 $\pm$ 2.85 & 78.7 $\pm$ 1.63 \\
StoryWriter & 78.2 $\pm$ 2.96 & 74.1 $\pm$ 3.69 \\
\textbf{\magnet} & \textbf{82.6} $\pm$ \textbf{1.60} & \textbf{79.6} $\pm$ \textbf{1.10} \\
\bottomrule
\end{tabular}

\caption{Rubric-based evaluation scores averaged across five generated stories.
Higher values indicate stronger narrative quality. Detailed story length
breakdowns can be found in~\ref{app:pairwise scores breakdown}.}
\label{tab:rubric_results}
\end{table}

\FloatBarrier

\begin{figure}[h]
    \centering

    \includegraphics[width=\linewidth]{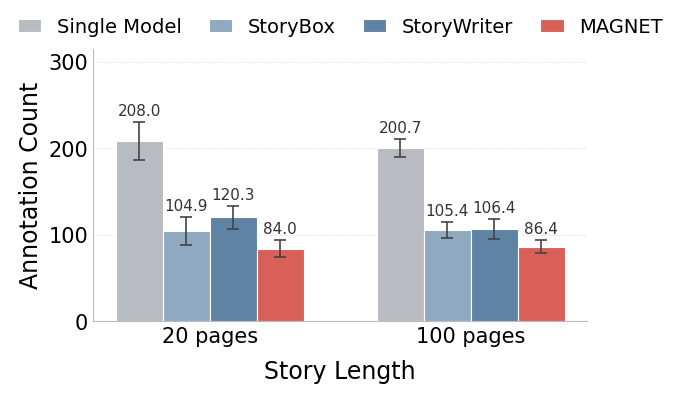}

    \textbf{(a)} Editorial Annotation Counts

    \vspace{0.8em}

    \includegraphics[width=\linewidth]{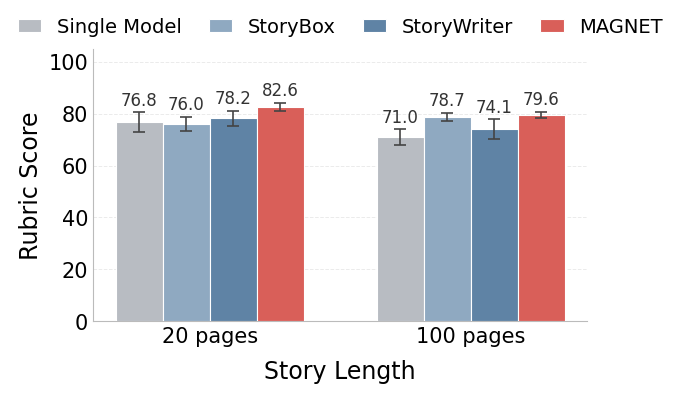}

    \textbf{(b)} Rubric-based Evaluation Scores

    \caption{
        Long-form story evaluation across generation methods at 20 and
        100 page story lengths. Error bars represent $\pm 1$ standard
        deviation. \textbf{(a)} Editorial annotation counts, where lower
        is better. \textbf{(b)} Rubric-based evaluation scores, where
        higher is better.
    }
    \label{fig:story_evaluation}
\end{figure}

\subsubsection{Human Validation}


To validate the quality of the LLM-generated editorial annotations and rubric assessments, we conducted human validation over a stratified sample of evaluations produced by both Haiku 4.5 and GPT-5.4 mini. We were blind to both the methodology (Magnet, Single Model, StoryBox, or StoryWriter) that produced each sampled story and the LLM evaluator (Haiku 4.5 or GPT-5.4 mini) that produced the annotation or rubric assessment under review. For each sampled annotation or rubric assessment, we independently read the relevant portion of the generated story and assessed whether the LLM evaluator's judgment was justified by the text.

For editorial annotations, we determined whether we agreed or disagreed with the LLM evaluator's assessment based on the story context. Agreement was defined as our determining that the identified issue was supported by the narrative and that the evaluator's underlying reasoning was reasonable. Disagreement was recorded when we determined that the annotation was not supported by the story or that the reasoning leading to the annotation was incorrect.

For rubric assessments, we similarly evaluated whether the reasoning underlying the assigned score was justified by the generated story. Because rubric scores are subjective and small numerical differences may reflect reasonable variation between evaluators, we did not seek to find an exact numerical match or impose an arbitrary point differential. Instead, we assessed whether the LLM evaluator's overall assessment and reasoning were consistent with the quality of the relevant story content.

We sampled 20 annotations per methodology (Magnet, Single Model, StoryBox, and StoryWriter) for each story length (20 and 100 pages), proportionally across story, chapter, and sentence level annotations. For rubric evaluation, we sampled the full-story score, two chapter-level scores, and three sentence-level scores for each methodology and story length.

Across the sampled evaluations, we agreed with 136 of 160 Haiku 4.5 annotations (85.0\%) and 132 of 160 GPT-5.4 mini annotations (82.5\%). For rubric assessments, we agreed with 41 of 48 Haiku 4.5 assessments (85.4\%) and 39 of 48 GPT-5.4 mini assessments (81.25\%). These results indicate that the LLM evaluators' editorial judgments and the reasoning supporting their rubric assessments were generally consistent with independent human assessment of the generated narratives.

\begin{table}[H]
\centering
\setlength{\tabcolsep}{7pt}
\renewcommand{\arraystretch}{0.9}

\begin{tabular}{@{}lccc@{}}
\toprule
\textbf{Model} & \textbf{Agreement} & \textbf{Total} & \textbf{\%} \\
\midrule
Haiku 4.5     & 136 & 160 & 85.0 \\
GPT 5.4 mini  & 132 & 160 & 82.5 \\
\bottomrule
\end{tabular}

\caption{Annotation Human Validation}
\label{tab:annotations_human_agreement}

\vspace{0.5em}

\begin{tabular}{@{}lccc@{}}
\toprule
\textbf{Model} & \textbf{Agreement} & \textbf{Total} & \textbf{\%} \\
\midrule
Haiku 4.5     & 41 & 48 & 85.4 \\
GPT 5.4 mini  & 39 & 48 & 81.25 \\
\bottomrule
\end{tabular}

\caption{Rubric Score Human Validation}
\label{tab:rubric_human_agreement}

\end{table}

\subsection{Ablation Analysis}
We perform an ablation using the hierarchical editorial evaluation framework on five 20-page stories to evaluate the contribution of individual framework components. Specifically, we evaluate variants of the system with the structured state updates removed, dynamic goal updates removed, and critic module removed (Table~\ref{tab:ablation}). For editorial annotations, removing the critic, structured state, and goal shifting significantly increased the number of annotations, with $p = 0.0216$, $p = 0.0001$, and $p = 0.0019$, respectively. The strong effects of removing structured world state tracking and goal shifting highlight the importance of dynamically adapting narrative state in long-form generation. Similarly, rubric-based scores decrease significantly for each ablation, with $p = 0.0111$, $p = 0.0023$, and $p = 0.0231$, respectively. These results demonstrate that all three components make statistically significant contributions to \textsc{Magnet's} narrative quality, with the full framework having the strongest performance across both evaluation metrics. 

\begin{table}[!htbp]
\centering
\footnotesize
\setlength{\tabcolsep}{3.5pt}
\renewcommand{\arraystretch}{0.95}

\begin{tabular}{@{}lcc@{}}
\toprule
\textbf{Config.} & \textbf{Annotations} $\downarrow$ & \textbf{Rubric} $\uparrow$ \\
\midrule
No Critic           & 109.0 $\pm$ 17.86 & 78.1 $\pm$ 1.18 \\
No Goal Shift       & 119.5 $\pm$ 12.34 & 78.4 $\pm$ 0.89 \\
No Structured State & 116.9 $\pm$ 10.76 & 76.4 $\pm$ 4.81 \\
\textbf{\magnet}    & \textbf{84.0 $\pm$ 9.64} & \textbf{82.6 $\pm$ 1.60} \\
\bottomrule
\end{tabular}

\caption{Ablation analysis on 20-page story generations averaged across
five generated stories. Lower editorial annotation counts and higher
rubric scores indicate better performance. Detailed category breakdowns
can be found in~\ref{app:detailed ablations}.}
\label{tab:ablation}
\end{table}

\FloatBarrier
\section{Conclusion}


In this work, we introduced \textsc{Magnet}, a multi-agent character-driven long-form narrative generation framework. Across all evaluations, \magnet significantly reduces editorial annotations and improves rubric-based scores compared to other story generation methods. Our ablation analysis further demonstrates that each component of \textsc{Magnet's} framework contributes significantly to its performance, highlighting the importance of iterative action generation, adaptive narrative plans, and structured states for story generation. Overall, this work provides a foundation for more controllable and structurally coherent long-form narrative generation. 


\section{Limitations}
While our work is effective at generating high-quality content that can be scaled to hundreds of pages, it remains computationally expensive due to repeated interaction between multiple generation modules. Thus, our editorial and scoring evaluations were performed on five \textsc{Magnet}-generated, StoryBox, StoryWriter, and single model stories. Additionally, \magnet depends on multiple closed-source models, which may introduce variability. Our evaluation pipeline, though human validated, depends on LLM-based judges, which may have subjective biases and reasoning.



\section{Ethical Considerations}

\textbf{Synthetic Content Misuses} Although \magnet improves long-form AI story generation, it could also be misused to create deceptive synthetic narratives. The increased narrative consistency and coherence may make the generated text appear more human-like, raising concerns regarding misinformation. Therefore, these AI-generated stories should be clearly disclosed, and appropriate safeguards and policies should be adhered to (\ref{app: licenses}).

\textbf{Human Creativity} Systems like \magnet may raise concerns about the impact of AI-generated narratives on human-created content. As models become increasingly capable of producing coherent long-form stories, the role of human writers could be greatly diminished, or it could lead to over-reliance on automated content generation. Therefore, AI-generated creative works should be used responsibly and appropriately attributed.

\textbf{Biases in Generated Content} Because narrative models learn from existing literary and web corpora, they risk reproducing cultural stereotypes, demographic biases, or toxic character tropes in generated story arcs. Therefore, deployments of this architecture should be monitored for harmful or discriminatory representations.


\newpage
\bibliography{custom}

@misc{yao2019planandwritebetterautomaticstorytelling,
      title={Plan-And-Write: Towards Better Automatic Storytelling}, 
      author={Lili Yao and Nanyun Peng and Ralph Weischedel and Kevin Knight and Dongyan Zhao and Rui Yan},
      year={2019},
      eprint={1811.05701},
      archivePrefix={arXiv},
      primaryClass={cs.CL},
      url={https://arxiv.org/abs/1811.05701}, 
}

@misc{lu2026assistantaxissituatingstabilizing,
      title={The Assistant Axis: Situating and Stabilizing the Default Persona of Language Models}, 
      author={Christina Lu and Jack Gallagher and Jonathan Michala and Kyle Fish and Jack Lindsey},
      year={2026},
      eprint={2601.10387},
      archivePrefix={arXiv},
      primaryClass={cs.CL},
      url={https://arxiv.org/abs/2601.10387}, 
}

@misc{shao2023characterllmtrainableagentroleplaying,
      title={Character-LLM: A Trainable Agent for Role-Playing}, 
      author={Yunfan Shao and Linyang Li and Junqi Dai and Xipeng Qiu},
      year={2023},
      eprint={2310.10158},
      archivePrefix={arXiv},
      primaryClass={cs.CL},
      url={https://arxiv.org/abs/2310.10158}, 
}

@misc{openai2024gpt4technicalreport,
      title={GPT-4 Technical Report}, 
      author={OpenAI},
      year={2024},
      eprint={2303.08774},
      archivePrefix={arXiv},
      primaryClass={cs.CL},
      url={https://arxiv.org/abs/2303.08774}, 
}

@misc{wang2024rolellmbenchmarkingelicitingenhancing,
      title={RoleLLM: Benchmarking, Eliciting, and Enhancing Role-Playing Abilities of Large Language Models}, 
      author={Zekun Moore Wang and Zhongyuan Peng and Haoran Que and Jiaheng Liu and Wangchunshu Zhou and Yuhan Wu and Hongcheng Guo and Ruitong Gan and Zehao Ni and Jian Yang and Man Zhang and Zhaoxiang Zhang and Wanli Ouyang and Ke Xu and Stephen W. Huang and Jie Fu and Junran Peng},
      year={2024},
      eprint={2310.00746},
      archivePrefix={arXiv},
      primaryClass={cs.CL},
      url={https://arxiv.org/abs/2310.00746}, 
}

@misc{gao2024agentscopeflexiblerobustmultiagent,
      title={AgentScope: A Flexible yet Robust Multi-Agent Platform}, 
      author={Dawei Gao and Zitao Li and Xuchen Pan and Weirui Kuang and Zhijian Ma and Bingchen Qian and Fei Wei and Wenhao Zhang and Yuexiang Xie and Daoyuan Chen and Liuyi Yao and Hongyi Peng and Zeyu Zhang and Lin Zhu and Chen Cheng and Hongzhu Shi and Yaliang Li and Bolin Ding and Jingren Zhou},
      year={2024},
      eprint={2402.14034},
      archivePrefix={arXiv},
      primaryClass={cs.MA},
      url={https://arxiv.org/abs/2402.14034}, 
}

@misc{park2023generativeagentsinteractivesimulacra,
      title={Generative Agents: Interactive Simulacra of Human Behavior}, 
      author={Joon Sung Park and Joseph C. O'Brien and Carrie J. Cai and Meredith Ringel Morris and Percy Liang and Michael S. Bernstein},
      year={2023},
      eprint={2304.03442},
      archivePrefix={arXiv},
      primaryClass={cs.HC},
      url={https://arxiv.org/abs/2304.03442}, 
}

@misc{li2026loststoriesconsistencybugs,
      title={Lost in Stories: Consistency Bugs in Long Story Generation by LLMs}, 
      author={Junjie Li and Xinrui Guo and Yuhao Wu and Roy Ka-Wei Lee and Hongzhi Li and Yutao Xie},
      year={2026},
      eprint={2603.05890},
      archivePrefix={arXiv},
      primaryClass={cs.CL},
      url={https://arxiv.org/abs/2603.05890}, 
}

@misc{Wang_2024,
      title={StoryVerse: Towards Co-authoring Dynamic Plot with LLM-based Character Simulation via Narrative Planning}, 
      author={Wang, Yi and Zhou, Qian and Ledo, David},
      year={2024},
      eprint={2405.13042},
      archivePrefix={arXiv},
      primaryClass={cs.AI},
      url={https://arxiv.org/abs/2405.13042}, 
}

@misc{huot2025agentsroomnarrativegeneration,
      title={Agents' Room: Narrative Generation through Multi-step Collaboration}, 
      author={Fantine Huot and Reinald Kim Amplayo and Jennimaria Palomaki and Alice Shoshana Jakobovits and Elizabeth Clark and Mirella Lapata},
      year={2025},
      eprint={2410.02603},
      archivePrefix={arXiv},
      primaryClass={cs.CL},
      url={https://arxiv.org/abs/2410.02603}, 
}

@misc{liu2026narrativetheorydrivenllmmethods,
      title={Narrative Theory-Driven LLM Methods for Automatic Story Generation and Understanding: A Survey}, 
      author={David Y. Liu and Aditya Joshi and Paul Dawson},
      year={2026},
      eprint={2602.15851},
      archivePrefix={arXiv},
      primaryClass={cs.CL},
      url={https://arxiv.org/abs/2602.15851}, 
}

@misc{surveyonllmsforstorygeneration, 
    title={A Survey on LLMs for Story Generation},
    author={Maria Teleki and Vedangi Bengali and Xiangjue Dong and Sai Tejas Janjur and Haoran Liu and Tian Liu and 
    Cong Wang and Ting Liu and Yin Zhang and Frank Shipman and James Caverlee},
    year={2025},
    url={https://aclanthology.org/2025.findings-emnlp.750.pdf},
}

@misc{lewis2021retrievalaugmentedgenerationknowledgeintensivenlp,
      title={Retrieval-Augmented Generation for Knowledge-Intensive NLP Tasks}, 
      author={Patrick Lewis and Ethan Perez and Aleksandra Piktus and Fabio Petroni and Vladimir Karpukhin and Naman Goyal and Heinrich Küttler and Mike Lewis and Wen-tau Yih and Tim Rocktäschel and Sebastian Riedel and Douwe Kiela},
      year={2021},
      eprint={2005.11401},
      archivePrefix={arXiv},
      primaryClass={cs.CL},
      url={https://arxiv.org/abs/2005.11401}, 
}

@misc{shinn2023reflexionlanguageagentsverbal,
      title={Reflexion: Language Agents with Verbal Reinforcement Learning}, 
      author={Noah Shinn and Federico Cassano and Edward Berman and Ashwin Gopinath and Karthik Narasimhan and Shunyu Yao},
      year={2023},
      eprint={2303.11366},
      archivePrefix={arXiv},
      primaryClass={cs.AI},
      url={https://arxiv.org/abs/2303.11366}, 
}

@misc{comanici2025gemini25pushingfrontier,
      title={Gemini 2.5: Pushing the Frontier with Advanced Reasoning, Multimodality, Long Context, and Next Generation Agentic Capabilities}, 
      author={Google DeepMind},
      year={2025},
      eprint={2507.06261},
      archivePrefix={arXiv},
      primaryClass={cs.CL},
      url={https://arxiv.org/abs/2507.06261}, 
}

@misc{opus4.7, 
    title={Claude Opus 4.7},
    author={Anthropic},
    year={2026},
    url={https://www.anthropic.com/claude/opus},
}

@misc{haiku4.5, 
    title={Claude Haiku 4.5},
    author={Anthropic},
    year={2026},
    url={https://www.anthropic.com/claude/haiku},
}

@misc{gemma4, 
    title={Gemma 4 model card},
    author={Google},
    year={2026},
    url={https://ai.google.dev/gemma/docs/core/model_card_4?},
}

@misc{gpt5.4mini, 
    title={GPT-5.4 mini Model},
    author={OpenAI},
    year={2026},
    url={https://developers.openai.com/api/docs/models/gpt-5.4-mini},
}

@misc{xia2025storywritermultiagentframeworklong,
      title={StoryWriter: A Multi-Agent Framework for Long Story Generation}, 
      author={Haotian Xia and Hao Peng and Yunjia Qi and Xiaozhi Wang and Bin Xu and Lei Hou and Juanzi Li},
      year={2025},
      eprint={2506.16445},
      archivePrefix={arXiv},
      primaryClass={cs.CL},
      url={https://arxiv.org/abs/2506.16445}, 
}

@misc{han2024ibsendirectoractoragentcollaboration,
      title={IBSEN: Director-Actor Agent Collaboration for Controllable and Interactive Drama Script Generation}, 
      author={Senyu Han and Lu Chen and Li-Min Lin and Zhengshan Xu and Kai Yu},
      year={2024},
      eprint={2407.01093},
      archivePrefix={arXiv},
      primaryClass={cs.CL},
      url={https://arxiv.org/abs/2407.01093}, 
}

@misc{packer2024memgptllmsoperatingsystems,
      title={MemGPT: Towards LLMs as Operating Systems}, 
      author={Charles Packer and Sarah Wooders and Kevin Lin and Vivian Fang and Shishir G. Patil and Ion Stoica and Joseph E. Gonzalez},
      year={2024},
      eprint={2310.08560},
      archivePrefix={arXiv},
      primaryClass={cs.AI},
      url={https://arxiv.org/abs/2310.08560}, 
}

@misc{borawski2026worldgenquestlinedependencydrivenprompt,
      title={From World-Gen to Quest-Line: A Dependency-Driven Prompt Pipeline for Coherent RPG Generation}, 
      author={Dominik Borawski and Marta Szulc and Robert Chudy and Małgorzata Giedrowicz and Piotr Mironowicz},
      year={2026},
      eprint={2604.25482},
      archivePrefix={arXiv},
      primaryClass={cs.CL},
      url={https://arxiv.org/abs/2604.25482}, 
}

@misc{wu2023autogenenablingnextgenllm,
      title={AutoGen: Enabling Next-Gen LLM Applications via Multi-Agent Conversation}, 
      author={Qingyun Wu and Gagan Bansal and Jieyu Zhang and Yiran Wu and Beibin Li and Erkang Zhu and Li Jiang and Xiaoyun Zhang and Shaokun Zhang and Jiale Liu and Ahmed Hassan Awadallah and Ryen W White and Doug Burger and Chi Wang},
      year={2023},
      eprint={2308.08155},
      archivePrefix={arXiv},
      primaryClass={cs.AI},
      url={https://arxiv.org/abs/2308.08155}, 
}

@misc{chi2025wowworldomniscientworld,
      title={WoW: Towards a World omniscient World model Through Embodied Interaction}, 
      author={Xiaowei Chi and Peidong Jia and Chun-Kai Fan and Xiaozhu Ju and Weishi Mi and Kevin Zhang and Zhiyuan Qin and Wanxin Tian and Kuangzhi Ge and Hao Li and Zezhong Qian and Anthony Chen and Qiang Zhou and Yueru Jia and Jiaming Liu and Yong Dai and Qingpo Wuwu and Chengyu Bai and Yu-Kai Wang and Ying Li and Lizhang Chen and Yong Bao and Zhiyuan Jiang and Jiacheng Zhu and Kai Tang and Ruichuan An and Yulin Luo and Qiuxuan Feng and Siyuan Zhou and Chi-min Chan and Chengkai Hou and Wei Xue and Sirui Han and Yike Guo and Shanghang Zhang and Jian Tang},
      year={2025},
      eprint={2509.22642},
      archivePrefix={arXiv},
      primaryClass={cs.RO},
      url={https://arxiv.org/abs/2509.22642}, 
}

@misc{ding2025understandingworldpredictingfuture,
      title={Understanding World or Predicting Future? A Comprehensive Survey of World Models}, 
      author={Jingtao Ding and Yunke Zhang and Yu Shang and Jie Feng and Yuheng Zhang and Zefang Zong and Yuan Yuan and Hongyuan Su and Nian Li and Jinghua Piao and Yucheng Deng and Nicholas Sukiennik and Chen Gao and Fengli Xu and Yong Li},
      year={2025},
      eprint={2411.14499},
      archivePrefix={arXiv},
      primaryClass={cs.CL},
      url={https://arxiv.org/abs/2411.14499}, 
}

@misc{li2026wordworldlargelanguage,
      title={From Word to World: Can Large Language Models be Implicit Text-based World Models?}, 
      author={Yixia Li and Hongru Wang and Jiahao Qiu and Zhenfei Yin and Dongdong Zhang and Cheng Qian and Zeping Li and Pony Ma and Guanhua Chen and Heng Ji},
      year={2026},
      eprint={2512.18832},
      archivePrefix={arXiv},
      primaryClass={cs.CL},
      url={https://arxiv.org/abs/2512.18832}, 
}

@misc{huang2026stateconsistencybehaviorconsistency,
      title={Beyond State Consistency: Behavior Consistency in Text-Based World Models}, 
      author={Youling Huang and Guanqiao Chen and Junchi Yao and Lu Wang and Fangkai Yang and Chao Du and ChenZhuo Zhao and Pu Zhao and Qingwei Lin and Saravan Rajmohan and Dongmei Zhang},
      year={2026},
      eprint={2604.13824},
      archivePrefix={arXiv},
      primaryClass={cs.LG},
      url={https://arxiv.org/abs/2604.13824}, 
}

@misc{cote2019textworldlearningenvironmenttextbased,
      title={TextWorld: A Learning Environment for Text-based Games}, 
      author={Marc-Alexandre Côté and Ákos Kádár and Xingdi Yuan and Ben Kybartas and Tavian Barnes and Emery Fine and James Moore and Ruo Yu Tao and Matthew Hausknecht and Layla El Asri and Mahmoud Adada and Wendy Tay and Adam Trischler},
      year={2019},
      eprint={1806.11532},
      archivePrefix={arXiv},
      primaryClass={cs.LG},
      url={https://arxiv.org/abs/1806.11532}, 
}

@misc{tian2026stagefullscreenplaybenchmarkreasoning,
      title={STAGE: A Full-Screenplay Benchmark for Reasoning over Evolving Storie}, 
      author={Qiuyu Tian and Zequn Liu and Yiding Li and Fengyi Chen and Zequn Liu and Youyong Kong and Fan Guo and Yuyao Li and Jinjing Shen and Zhijing Xie and Yiyun Luo and Xin Zhang and Yingce Xia},
      year={2026},
      eprint={2601.08510},
      archivePrefix={arXiv},
      primaryClass={cs.CL},
      url={https://arxiv.org/abs/2601.08510}, 
}

@misc{wu2025longevalcomprehensiveanalysislongtext,
      title={LongEval: A Comprehensive Analysis of Long-Text Generation Through a Plan-based Paradigm}, 
      author={Siwei Wu and Yizhi Li and Xingwei Qu and Rishi Ravikumar and Yucheng Li and Tyler Loakman and Shanghaoran Quan and Xiaoyong Wei and Riza Batista-Navarro and Chenghua Lin},
      year={2025},
      eprint={2502.19103},
      archivePrefix={arXiv},
      primaryClass={cs.CL},
      url={https://arxiv.org/abs/2502.19103}, 
}

@misc{zheng2023judgingllmasajudgemtbenchchatbot,
      title={Judging LLM-as-a-Judge with MT-Bench and Chatbot Arena}, 
      author={Lianmin Zheng and Wei-Lin Chiang and Ying Sheng and Siyuan Zhuang and Zhanghao Wu and Yonghao Zhuang and Zi Lin and Zhuohan Li and Dacheng Li and Eric P. Xing and Hao Zhang and Joseph E. Gonzalez and Ion Stoica},
      year={2023},
      eprint={2306.05685},
      archivePrefix={arXiv},
      primaryClass={cs.CL},
      url={https://arxiv.org/abs/2306.05685}, 
}

@misc{liu2023gevalnlgevaluationusing,
      title={G-Eval: NLG Evaluation using GPT-4 with Better Human Alignment}, 
      author={Yang Liu and Dan Iter and Yichong Xu and Shuohang Wang and Ruochen Xu and Chenguang Zhu},
      year={2023},
      eprint={2303.16634},
      archivePrefix={arXiv},
      primaryClass={cs.CL},
      url={https://arxiv.org/abs/2303.16634}, 
}

@misc{rao2026autorubricunifyingrubricbasedllm,
      title={Autorubric: Unifying Rubric-based LLM Evaluation}, 
      author={Delip Rao and Chris Callison-Burch},
      year={2026},
      eprint={2603.00077},
      archivePrefix={arXiv},
      primaryClass={cs.CL},
      url={https://arxiv.org/abs/2603.00077}, 
}

@misc{kim2024prometheusinducingfinegrainedevaluation,
      title={Prometheus: Inducing Fine-grained Evaluation Capability in Language Models}, 
      author={Seungone Kim and Jamin Shin and Yejin Cho and Joel Jang and Shayne Longpre and Hwaran Lee and Sangdoo Yun and Seongjin Shin and Sungdong Kim and James Thorne and Minjoon Seo},
      year={2024},
      eprint={2310.08491},
      archivePrefix={arXiv},
      primaryClass={cs.CL},
      url={https://arxiv.org/abs/2310.08491}, 
}

@misc{gu2025surveyllmasajudge,
      title={A Survey on LLM-as-a-Judge}, 
      author={Jiawei Gu and Xuhui Jiang and Zhichao Shi and Hexiang Tan and Xuehao Zhai and Chengjin Xu and Wei Li and Yinghan Shen and Shengjie Ma and Honghao Liu and Saizhuo Wang and Kun Zhang and Yuanzhuo Wang and Wen Gao and Lionel Ni and Jian Guo},
      year={2025},
      eprint={2411.15594},
      archivePrefix={arXiv},
      primaryClass={cs.CL},
      url={https://arxiv.org/abs/2411.15594}, 
}

@misc{chen2024llmstateopenworldstate,
      title={LLM-State: Open World State Representation for Long-horizon Task Planning with Large Language Model}, 
      author={Siwei Chen and Anxing Xiao and David Hsu},
      year={2024},
      eprint={2311.17406},
      archivePrefix={arXiv},
      primaryClass={cs.RO},
      url={https://arxiv.org/abs/2311.17406}, 
}

@misc{que2024hellobenchevaluatinglongtext,
      title={HelloBench: Evaluating Long Text Generation Capabilities of Large Language Models}, 
      author={Haoran Que and Feiyu Duan and Liqun He and Yutao Mou and Wangchunshu Zhou and Jiaheng Liu and Wenge Rong and Zekun Moore Wang and Jian Yang and Ge Zhang and Junran Peng and Zhaoxiang Zhang and Songyang Zhang and Kai Chen},
      year={2024},
      eprint={2409.16191},
      archivePrefix={arXiv},
      primaryClass={cs.CL},
      url={https://arxiv.org/abs/2409.16191}, 
}

@article{Chen_2026,
   title={StoryBox: Collaborative Multi-Agent Simulation for Hybrid Bottom-Up Long-Form Story Generation Using Large Language Models},
   volume={40},
   ISSN={2159-5399},
   url={http://dx.doi.org/10.1609/aaai.v40i36.40288},
   DOI={10.1609/aaai.v40i36.40288},
   number={36},
   journal={Proceedings of the AAAI Conference on Artificial Intelligence},
   publisher={Association for the Advancement of Artificial Intelligence (AAAI)},
   author={Chen, Zehao and Pan, Rong and Li, Haoran},
   year={2026},
   month=Mar, pages={30359–30367} 
}

\appendix
\section{Appendix}
\label{sec:appendix}

\FloatBarrier
\subsection{Detailed Experimental Setup} \label{app:expsetup}

\textbf{Models} The critic module uses a temperature of 0.1. The narrator uses a temperature of 0.35, and the goal generator uses a temperature of 0.1. The character agent uses a temperature of 0.2. The editorial and rubric scoring evaluation uses a temperature of 0.1. The baseline model uses a temperature of 0.35. The aforementioned temperatures were derived from experimentally testing each model's temperature hyperparameter to optimize for its generation performance. 

\textbf{Story Genres} We evaluate five story genres: Mystery (Missing Codicil), Thriller (Museum), Action (Hospital), Social Drama (Warehouse), and Workplace Drama (Orchestra).

\textbf{StoryBox} For evaluating StoryBox, we used gpt-5.4-mini for running the character interactions and gpt-5.4 for the prose writing.

\textbf{StoryWriyer} For evaluating StoryWriter, we used gpt-5.4-mini for generating the outline and gpt-5.4 for the prose writing.

\textbf{Computational Costs}
In total, we spent \$588.42 on API usage for closed-source models, including \$144.77 on Gemini, \$110.68 on OpenAI, and \$322.97 on Anthropic. Generating a 20 page story with \magnet costs approximately \$6 in API usage, while generating a 100 page story costs approximately \$30. All experiments were ran on an NVIDIA RTX 5090 GPU. 

\FloatBarrier


\subsection{Detailed Editor Annotation Counts} \label{app:editor annotation breakdown}

\begin{table}[H]
\centering
\footnotesize
\setlength{\tabcolsep}{2.5pt}
\renewcommand{\arraystretch}{1.05}

\begin{tabular}{@{}llccc@{}}
\toprule
\textbf{Method} & \textbf{Length} &
\textbf{Story} $\downarrow$ &
\textbf{Chapter} $\downarrow$ &
\textbf{Sentence} $\downarrow$ \\
\midrule

\multirow{2}{*}{Base}
& 20 p.  & 29.2 $\pm$ 4.45 & 155.1 $\pm$ 18.51 & 23.7 $\pm$ 7.27 \\
& 100 p. & 37.2 $\pm$ 3.65 & 141.7 $\pm$ 13.84 & 21.8 $\pm$ 3.72 \\
\midrule

\multirow{2}{*}{StoryBox}
& 20 p.  & 28.9 $\pm$ 4.60 & 64.3 $\pm$ 13.34 & 11.7 $\pm$ 1.44 \\
& 100 p. & 28.8 $\pm$ 4.88 & 61.5 $\pm$ 8.56 & 15.1 $\pm$ 1.52 \\
\midrule

\multirow{2}{*}{StoryWriter}
& 20 p.  & 29.4 $\pm$ 3.07 & 78.6 $\pm$ 13.96 & 12.3 $\pm$ 2.31 \\
& 100 p. & 29.1 $\pm$ 8.52 & 62.5 $\pm$ 7.85 & 14.8 $\pm$ 1.30 \\
\midrule

\multirow{2}{*}{\magnet}
& 20 p.  & 16.5 $\pm$ 5.89 & 48.2 $\pm$ 7.29 & 19.3 $\pm$ 0.84 \\
& 100 p. & 14.2 $\pm$ 6.23 & 53.5 $\pm$ 11.71 & 18.7 $\pm$ 3.38 \\

\bottomrule
\end{tabular}

\caption{Editorial annotation counts across story hierarchies. Results are
reported as mean $\pm$ standard deviation. Lower is better.}
\label{tab:annotation_scores_appendix}
\end{table}

\subsection{Detailed Rubric Scores} 
\label{app:pairwise scores breakdown}

\begin{table}[H]
\centering
\footnotesize
\setlength{\tabcolsep}{2.5pt}
\renewcommand{\arraystretch}{1.05}

\begin{tabular}{@{}llccc@{}}
\toprule
\textbf{Method} & \textbf{Length} &
\textbf{Story} $\uparrow$ &
\textbf{Chapter} $\uparrow$ &
\textbf{Sentence} $\uparrow$ \\
\midrule

\multirow{2}{*}{Base}
& 20 p.  & 87.1 $\pm$ 2.75 & 79.2 $\pm$ 4.03 & 64.0 $\pm$ 7.59 \\
& 100 p. & 81.5 $\pm$ 15.60 & 71.0 $\pm$ 7.06 & 60.6 $\pm$ 3.04 \\
\midrule

\multirow{2}{*}{StoryBox}
& 20 p.  & 84.4 $\pm$ 1.39 & 82.0 $\pm$ 6.80 & 61.7 $\pm$ 5.00 \\
& 100 p. & 86.2 $\pm$ 1.89 & 83.0 $\pm$ 4.32 & 66.9 $\pm$ 2.82 \\
\midrule

\multirow{2}{*}{StoryWriter}
& 20 p.  & 82.5 $\pm$ 3.16 & 84.3 $\pm$ 1.23 & 67.6 $\pm$ 6.51 \\
& 100 p. & 77.3 $\pm$ 12.82 & 82.1 $\pm$ 6.07 & 62.8 $\pm$ 7.64 \\
\midrule

\multirow{2}{*}{MAGNET}
& 20 p.  & 86.2 $\pm$ 4.02 & 81.7 $\pm$ 3.94 & 79.9 $\pm$ 1.90 \\
& 100 p. & 80.0 $\pm$ 2.72 & 82.4 $\pm$ 3.63 & 76.6 $\pm$ 3.02 \\

\bottomrule
\end{tabular}

\caption{Rubric-based evaluation scores across story hierarchies. Results
are reported as mean $\pm$ standard deviation. Higher is better.}
\label{tab:rubric_scores_appendix}
\end{table}

\subsection{Detailed Ablation Annotation Counts} \label{app:detailed ablations}
\begin{table}[h]
\centering
\footnotesize
\setlength{\tabcolsep}{2.5pt}
\renewcommand{\arraystretch}{1.05}

\begin{tabular}{@{}lccc@{}}
\toprule
\textbf{Ablation} 
& \textbf{Story} $\downarrow$ 
& \textbf{Chapter} $\downarrow$ 
& \textbf{Sentence} $\downarrow$ \\
\midrule

No Critic 
& 17.0 $\pm$ 6.17 
& 72.3 $\pm$ 13.10 
& 19.7 $\pm$ 2.11 \\

No Goal Shift 
& 22.2 $\pm$ 8.85 
& 77.3 $\pm$ 14.99 
& 20.0 $\pm$ 3.54 \\

No Structured State 
& 20.5 $\pm$ 10.00 
& 78.0 $\pm$ 5.16 
& 18.4 $\pm$ 1.56 \\

\bottomrule
\end{tabular}

\caption{Detailed editorial annotation counts for ablations of MAGNET.
Results are reported as mean $\pm$ standard deviation across generated
stories. Lower is better.}
\label{tab:ablation_annotation_breakdown}
\end{table}

\begin{table}[h]
\centering
\footnotesize
\setlength{\tabcolsep}{2.5pt}
\renewcommand{\arraystretch}{1.05}

\begin{tabular}{@{}lccc@{}}
\toprule
\textbf{Ablation}
& \textbf{Story} $\uparrow$
& \textbf{Chapter} $\uparrow$
& \textbf{Sentence} $\uparrow$ \\
\midrule

No Critic 
& 81.3 $\pm$ 6.06 
& 78.1 $\pm$ 8.32 
& 74.8 $\pm$ 2.04 \\

No Goal Shift 
& 83.9 $\pm$ 6.57 
& 74.7 $\pm$ 8.44 
& 76.5 $\pm$ 4.07 \\

No Structured State
& 83.9 $\pm$ 4.95 
& 73.3 $\pm$ 7.78 
& 72.1 $\pm$ 7.80 \\

\bottomrule
\end{tabular}

\caption{Detailed rubric-based evaluation scores for ablations of MAGNET.
Results are reported as mean $\pm$ standard deviation across generated
stories. Higher is better.}
\label{tab:ablation_rubric_breakdown}
\end{table}

\subsection{Licenses} \label{app: licenses}

Code and anonymized repositories associated with this work are released under the MIT License:
{\small
\begin{itemize}
    \item \magnet: \url{https://anonymous.4open.science/r/MAGNET-ED2F/README.md}
\end{itemize}
}

Gemma 4 is distributed under the Apache License 2.0
(\url{https://ai.google.dev/gemma/docs/core}).

StoryBox's codebase is distributed under the MIT License (\url{https://github.com/amcghm/StoryBox}).

StoryWriter's codebase is distributed under the MIT License (\url{https://github.com/THU-KEG/StoryWriter}).

All models and code are used in accordance with their respective licenses or terms of service.

\end{document}